# Modular AUV System for Sea Water Quality Monitoring and Management

Mike Eichhorn, Ralf Taubert, Christoph Ament
Institute for Automation and Systems Engineering
Ilmenau University of Technology
98684 Ilmenau, Germany
Email: mike.eichhorn@tu-ilmenau.de

Marco Jacobi, Torsten Pfuetzenreuter
Branch Advanced System Technology (AST)
Fraunhofer IOSB
98693 Ilmenau, Germany
Email: marco.jacobi@iosb-ast.fraunhofer.de

*Abstract* — The sustained and cost-effective monitoring of the water quality within European coastal areas is of growing importance in view of the upcoming European marine and maritime directives, i.e. the increased industrial use of the marine environment. Such monitoring needs mechanisms/systems to detect the water quality in a large sea area at different depths in real time.

This paper presents a system for the automated detection and analysis of water quality parameters using an autonomous underwater vehicle. The analysis of discharge of nitrate into Norwegian fjords near aqua farms is one of the main application fields of this AUV system. As carrier platform the AUV "CWolf" from the Fraunhofer IOSB-AST will be used, which is perfectly suited through its modular payload concept. The mission task and the integration of the payload unit which includes the sensor module, the scientific and measurement computer in the AUV carrier platform will be described. Few practice oriented information about the software and interface concept, the function of the several software modules and the test platform with the several test levels to test every module will be discussed.

I. INTRODUCTION

Guaranteeing the cleanliness of oceans is of crucial importance, and efforts should be made to preserve it as a sustainable habitat. Inshore waters are expected to be used considerably more widely for agricultural activities or for raw materials extraction in the future. Examples of this include fish farming in aquacultures, mining, or offshore oil production. Methods for waste avoidance must be developed, which minimize the input of pollutants into waters. Indeed, the need for such measures, and accordingly the improvement of environmental quality, has become of increasing interest by European agencies. The Water Information System for Europe (WISE) demands, for example, detailed information about the status of the water quality of coastal ocean waters [1]. Such requirements are only accomplishable with a frequent logging of the water quality and their biological cause variables.

The current state of technology is the use of specific research vessels to conduct such investigations. This requires an extremely high resource management on the one hand, and a comparatively long preparation period for their use on the other.

The "FerryBox" project shows one possible approach in the field of automatic water quality measurement [2]. The idea is to use existing platforms (ferry, containership, etc.) which are cost-oriented and without additional technical expenditure. However, this approach can be used only on the routes of the mentioned carrier vehicles and is therefore inflexible. Through the miniaturization of such a measurement system and its combination with the flexibility of an Autonomous Underwater Vehicle (AUV), the base for an automatic and closely meshed monitoring of inshore waters, fjords and inland waters will be created.

One possibility is the usage of AUVs called gliders [3, 4]. These gliders have a low cruising speed (0.2 to 0.4 m s-1) for long operational periods up to 30 days with low energy consumption achieved by the passive drive concept. Since the payload capacity is limited, there are restrictions in sensor weight and volume as well as in energy consumption. Alternatively, this measurement system could be deployed on buoys, as described in [5], is a possible solution. This system is limited however, since measurement data can only be collected from one single position.

This paper presents a solution using a mission proven AUV as a carrier platform to implement a miniaturized measurement system to analyze water quality. The presented system was developed within the research project SALMON (Sea Water Quality Monitoring and Management). On this project German (Ilmenau University of Technology, Fraunhofer IOSB-AST, 4H-JENA engineering GmbH), Norwegian (Havforskningsinstituttet, Institute of Marine Research) and Danish (Mads Clausen Institute) companies and research institutions work on a systematic solution for automatic water monitoring. One goal is to show, by example, the import of nutritive substances in Norwegian aqua farms located in fjords. These fjords can be monitored with this system and can be refined in order to reduce outside influences on nature.

The concept presented illustrates how the measurement system can be integrated easily into the vehicle with a minimum of hard- and software technical interfaces. Moreover, several software technical details to achieve this goal were presented. This paper focusses on presenting the test system using MATLAB/Simulink [6, 7]. This allows for testing of individual C++ modules with easy, effective observation of the interaction between modules, and rapid provision of test results of complex test scenarios.

This work was supported by the European Regional Development Fund (ERDF) of the European Union via the Thuringian Coordination Office TNA #TNA VIII-1/2011

## II. MISSION TASK

To detect the distribution of nitrates around the aqua farm, a mission plan was created, that is based horizontally on a meander with saw tooth shaped dive profiles. Thus, it is possible to measure the nitrate concentration at several depths and to support navigation of the AUV with actual GPS positions during the surface phases.

### A. Definition of the Mission Plan

To create a mission plan a menu-guided planning system will be used. The user will be presented step by step with information and dialogs, which are necessary to solve the actual planning task. The typical planning sequence includes three stages: 1) defining the sea chart/area of interest; 2) selection of the vehicle; and 3) build a plan using defined mission elements. Fig. 1 shows the possible mission elements (vehicle primitives). Available elements include an initial and a final element (to define the start and goal position of the mission), the three base maneuver elements, (waypoint, line and arc) and the complex mission element meander. The configuration of several elements can be started by choosing the respective element in the lower half of the window. The meander is specified with the following parameters: start position $x_{meander}$, $y_{meander}$, depth $z_{max}$, rotation $\theta_{meander}$, leg length $l_{leg}$, distance between two legs $d_{leg}$ and the number of legs $n_{legs}$ (see Fig. 2). The generated mission plan will be stored in a text file. At the beginning of a mission the *Maneuver Processor* loads and parses this text file in a map structure to use it during the mission.

### B. Define a Waypoint List used in Trajectory Planning

The control of the vehicle takes place with a state based tracking controller [8]. This controller requires waypoints to create a reference trajectory using a cubic spline interpolation. In order for the reference trajectory to correspond as closely as possible with the defined meander which consists of lines and arcs, characteristic waypoints have to be defined. The requirements for such waypoints are:

- smoothness conditions on the reference trajectory, needed within the controller design,
- good reproduction of the saw tooth shaped dive profiles and
- a small number of waypoints.

Fig. 2 shows several parameters to define a waypoint list according to the route of the horizontal path and the associated vertical dive profiles. The angle $\alpha_{arc}$ will be used for the reproduction of the arc element.

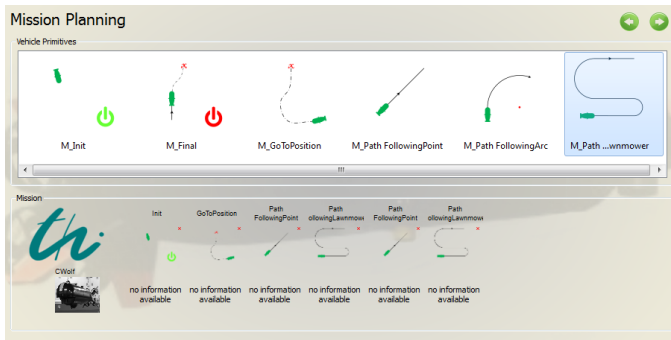

Fig. 1. User menu for element-based planning of a mission

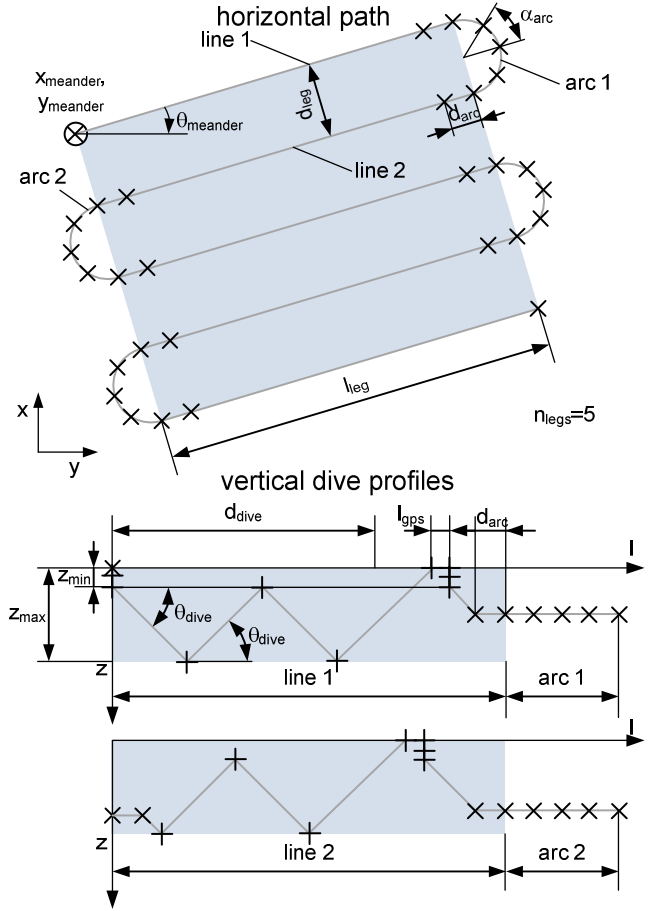

Fig. 2. Parameters used to define the waypoint list

This also defines the angle differences between adjacent waypoints on the arc. To generate a smooth transition between a line segment and an arc, a way-point will be positioned before the start and behind the end position of the arc at a distance of $d_{arc}$. In the figure above these waypoints are signed with a "x" mark.

The depth information for the defined waypoints will be generated from the required dive profile. The chosen dive profile has a saw tooth shape similar to a Slocum Glider dive profile [9]. This allows the recording of measurement data at every depth in the least amount of time using a minimal number on course changes. Therefore, the area of interest can be recorded in a minimal duration with small energy consumption by the AUV. To support the navigation with actual GPS positions, a surface drive is required when the distance of the underwater drive covered reaches $d_{dive}$. In the figure above this occurs on each leg element. It is also possible that the surface maneuver takes place after n legs. This is dependent on the leg length $l_{leg}$ and the submerge and emerge angle $\alpha_{dive}$. The track length of the surface drive is defined by the parameter $l_{gps}$. Due to the construction properties of the AUV a diagonal submerge maneuver is impossible (see section III.A). At the beginning of the submerge maneuver the AUV uses the thruster. In a depth of $z_{min}$ the vehicle uses the propellers and starts with the diagonal maneuver. This depth will also be used for the initiation of a submerge maneuver after an emerge maneuver. Thus, the defined waypoints have a "+" mark in the figure.

## III. HARDWARE

The AUV base platform includes the physical vehicle with its actuators, navigation and surveillance sensors, communication units and a Control Computer (CC). It provides an interface to receive the navigation data, the vehicle status and error messages and control several actuators. The vehicle guidance occurs on a Scientific Computer (SC), which is located in the payload unit. A Measurement Computer (MC) for the nitrate measuring management is also located there. This concept allows an easy integration of the application task with a minimum of hard- and software technical interfaces.

### A. AUV Base Platform

The AUV CWolf (see Fig. 3) is able to conduct autonomous missions with very different characteristics using *ConSys* [10]. For the SALMON project, a track controller is required that runs on the SC in the payload unit. The vehicle is based on the SeaWolf ROV from the German company ATLAS ELEKTRONIK, Bremen. It is equipped with typical navigation sensors (pressure, FOG, DVL, GPS) and uses an acoustic modem with the USBL localization option to exchange data with a control station. A forward-looking sonar is used for obstacle detection and avoidance.

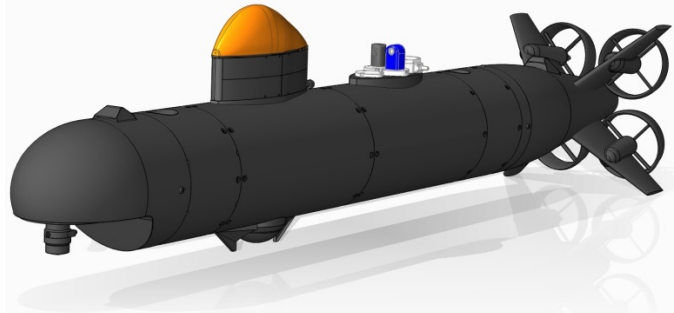

Fig. 3. AUV CWolf

The CWolf is equipped with four propulsion units to advance the vehicle, as well as to control depth, yaw, and pitch. Two vertical thrusters at bow and stern are used for depth and pitch control during low speed operation. The most important technical data of the AUV are shown in TABLE I.

TABLE I. TECHNICAL DATA OF CWOLF AUV

| Parameter | Value |
|---|---|
| Length | 2,20 m |
| Diameter | 0,30 m |
| Weight in air | 135 kg |
| Max. speed | 6 kn |
| Endurance at 3 kn | 3 h |
| Payload | 15 kg |

### B. Payload Unit

#### 1) Computer Units

The Scientific and Measurement Computer uses a Pokini Z550 platform. According to the manufacturer, this computer is the smallest fan-less computer in the world [11] with a size of 115 x 27 x 101 mm. This computer has an Intel Atom Processor Z550, 2 GB RAM and a Solid State Drive Intel 320 Series SSD 120GB. The operating system is Windows 7 32 bit. Fig. 4 shows the backside of the computer with the available interfaces.

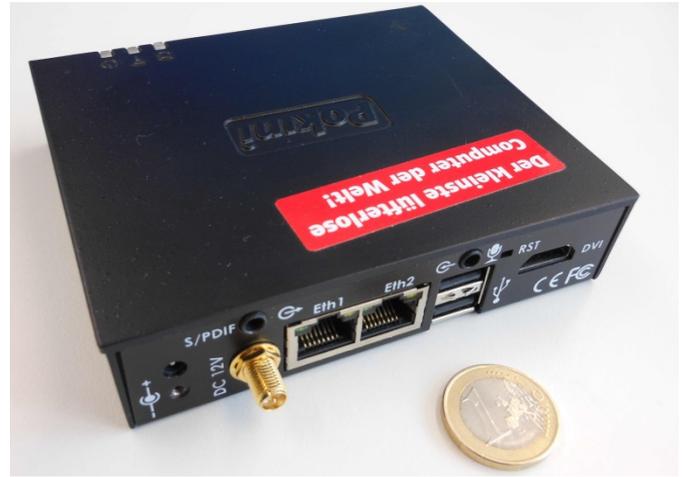

Fig. 4. Computer Pokini Z550

#### Sensor System

The project partner 4H JENA engineering GmbH [12] developed a miniaturized sensor system for the payload unit. TABLE II shows the main parameters of the system. The main components and sensors are shown in Fig. 5.

TABLE II. PARAMETERS OF THE SENSOR SYSTEM

| Parameter | Measuring Range/Value |
|---|---|
| Sodium nitrate (NaNO3) | 0-1000 µg/l |
| Oxygen concentration ($O_2$) | 0-500 µmol/l |
| Conductivity σ | 0-75 mS/cm |
| Temperature T | -5-40 °C |
| Measurement cycle | 1 s ($O_2$, σ)/ 5-10 s (NaNO3) |
| Power Supply (Computer) | 12 V |
| Power Supply (Sensors) | 19-25 V |
| Energy Consumption | 12 W (12 V)/10 W (22 V) |

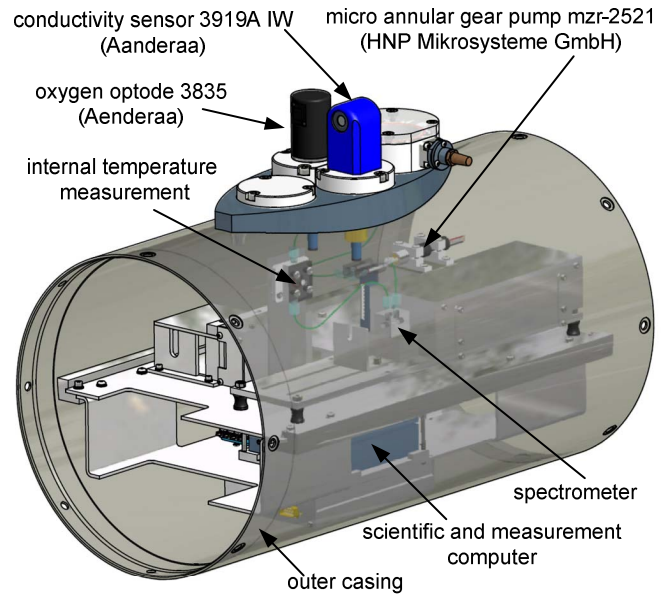

Fig. 5. CAD drawing of the Sensor System (4H JENA engineering GmbH)

## IV. SOFTWARE TECHNICAL DETAILS

### A. Communication Concept

For communication between the programs on the computers, UDP (User Data Protocol) will be used. This allows operating system and compiler independency of the programs. The socket classes on the scientific computer use the Asio library, which is a component of the Boost library [13]. Asio is a cross-platform C++ library for asynchronous data processing and for network programming. Fig. 6 shows the interfaces between the programs and the defined message telegrams for exchange of information/data and for delivery of control commands.

The telegram *LLC_Command* is needed to activate the vehicle control from the Scientific Computer (SC). Therefore, it can be switched between a controlled mode, a direct mode, or no-control mode, when the SC finishes the vehicle control. In controlled mode, the LLC requires a heading, depth and speed command. In the SALMON project, direct mode will be used whereby the SC controls the vehicle propulsion engines directly. It has to send *LLC_Setpoints_CWolf* telegrams which include the PWM-values for each single motor.

The Control Computer distributes the actual navigation data, which is used from the SC and MC to handle the mission, to control the vehicle and to record the water quality measurements. The *NAV_Data* telegram includes the actual vehicle location (latitude, longitude and depth values), the vehicle orientation, speed, and height over ground values, when available. The Low Level Controller (LLC) of the CC communicates between the motor hardware and the CC-software. It receives the *LLC_Setpoint_CWolf* and *LLC_Setpoint* telegrams and sends *LLC_Status*, with the actual motor status, and *LLC_Error* telegrams, when an error has occurred. Each telegram has to be identified. The unique identification of telegrams occurs with a header, which contains a message id and the message-payload length.

### B. Program Interface Structure

The interfaces of the programs (*Mission Management*, *Maneuver Processor* and *Autopilot*) which run on the Scientific Computer are based on a bridge design pattern [14]. In this pattern, there exist abstract C++ classes which define the public interfaces. That means every program has an abstract class *InputProcessor_C* and *OutputProcessor_C*. The derivations of these classes include the code of the interface type being used. Hence, the use of another interface type (Serial port, TCP/IP, UDP, CORBA, …) is easy to handle. The actual code needs not be modified in such a case. Another reason for such an interface design results from the test platform in Simulink (see section V.A). Every program code is implemented in an S-Function, where the communication between the programs works with S-Function methods and S-Function data structures. These program parts are included in the interface classes. Fig. 7 shows this concept by means of the autopilot system.

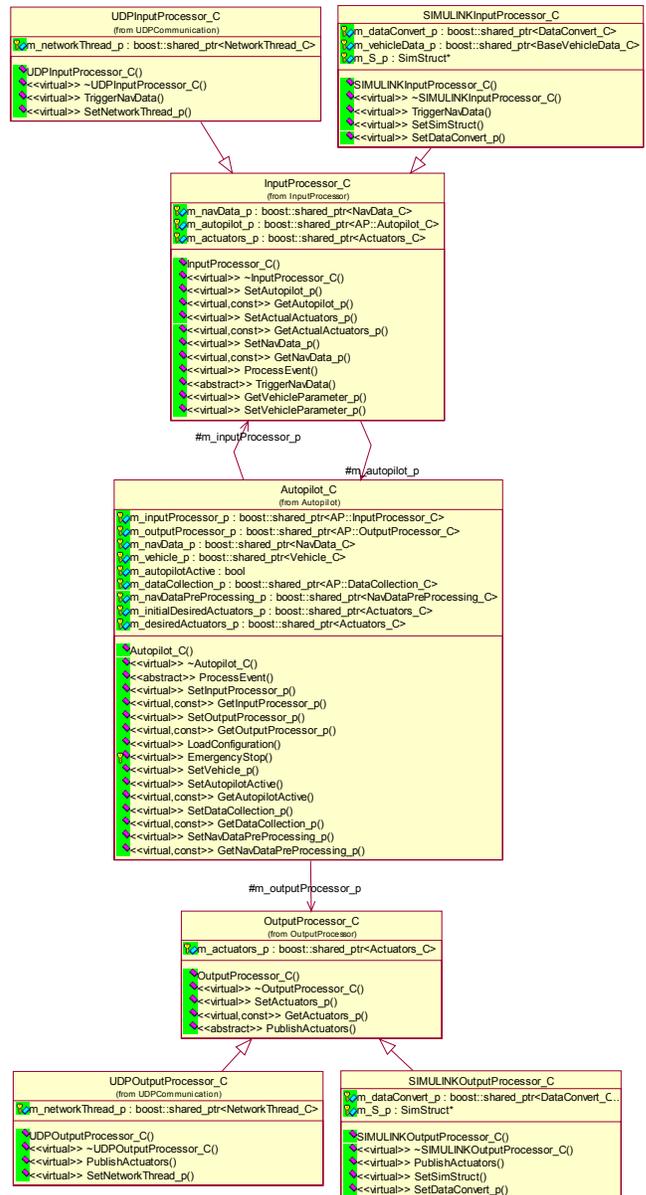

Fig. 7. Class diagram of the Autopilot System

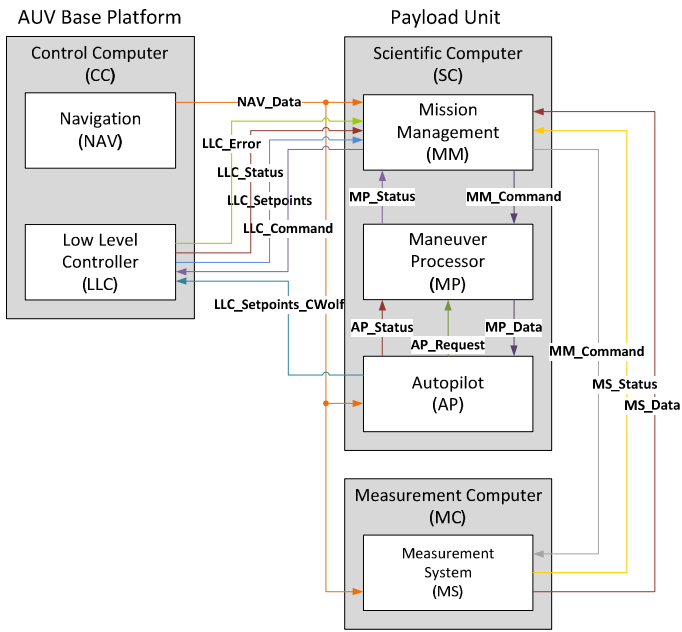

Fig. 6. Interfaces and message telegrams between the computers

## C. Model-Based Design of the Autopilot

For ambitious highly-complex projects like the development of an AUV autopilot system [8], short design cycles are necessary. Often false or incomplete implementations of the requirements cause errors, which may only be discovered late in the development process. Model-Based Design (MBD) prevents time delays and cost overruns, which result from late error fixing, with continuous verification and validation in every development phase [15]. This design concept allows an easy adaption/modification of the autopilot system in the start-up period, where the autopilot has to control the real vehicle during sea trials.

After capturing the requirements, MATLAB/Simulink [6, 7] are used for the design and the modeling of the autopilot system. The integrated subsystems consist of classic block diagrams and textual programming language like M-functions or S-functions.

Implementation of the *Autopilot* is the next step in the development process. MATLAB/Simulink can create C++ code automatically using the Simulink Coder™ [16] (formerly Real-Time Workshop®). The generated header- and source-files reproduce the whole functionality of the *Autopilot* in two main functions (*initialize* and *step*). After integration in the *Autopilot_C* class and the definition of the input/output interfaces, a library is created.

This library can be used to build a compiled C-Mex-S-Function for MATLAB/Simulink. The masked S-Function replaces the translated subsystem in the simulation and validates the functional behavior of the coded Autopilot. At this point a comparison with the designed subsystem saves the right translation of the Simulink Coder. To assure the real-time behavior, an executable file is transferred onto the target platform PC (Pokini Z550). In the HIL (Hardware-in-the-loop) mode, it is possible to simulate the *Autopilot* directly on the embedded system.

In summary, the Model-Based Design, in combination with automatic coder tools, provides a powerful instrument for generating executable source code without manually typing any C++ lines. The usage of MATLAB scripts, which defines all options and steps during the code generation, allows a fast implementation on the target system after changing the Simulink model in the system design phase. Finally, when the autopilot model fulfills the requirements, the source code will be optimized manually.

## V. TEST SYSTEM

To achieve high software quality and safe algorithms all programs have to be tested. The test system presented in this section relates mainly to the newly developed software in the Scientific Computer. These are the *Mission Management*, the *Maneuver Processor* and the *Autopilot*. Testing of the modules takes place in several phases.

A simulation environment using MATLAB/Simulink was developed for the tests as well as for the development of the modules. For the module code to run in Simulink Wrapper S-Functions have to be written (A detailed description about this procedure is presented in section V.A). The simulation environment allows for:

- Effect troubleshooting by attaching the Visual Studio debugger to the MATLAB/Simulink process that is running outside of Visual Studio. (Every module has a Visual Studio project to build a C MEX S-Function using the module C++ code. This enables the monitoring of variables and the definition of breakpoints inside the module code in Debug-Mode.)

- Single step mode during the debugging.

- Processing of the modules is many times faster than in real time (This is an interesting fact in case of running a complex test scenario after program changes).

- A separate test of the modules.

- Definition of complex test sequence scripts in MATLAB.

- Easy analysis of the tests and the display of the results in MATLAB.

- Optimization of control parameters in the *Autopilot* according to user-defined requirements.

At this point, it should be mentioned that only the *Autopilot* module was designed in MATLAB/Simulink (see section IV.C). The other modules do not use any MATLAB functionality. For programs which use complex data structures or which require additional libraries the C++ code generation via MBD cannot be recommended. Furthermore, the code generated by MBD is not transparent and optimal.

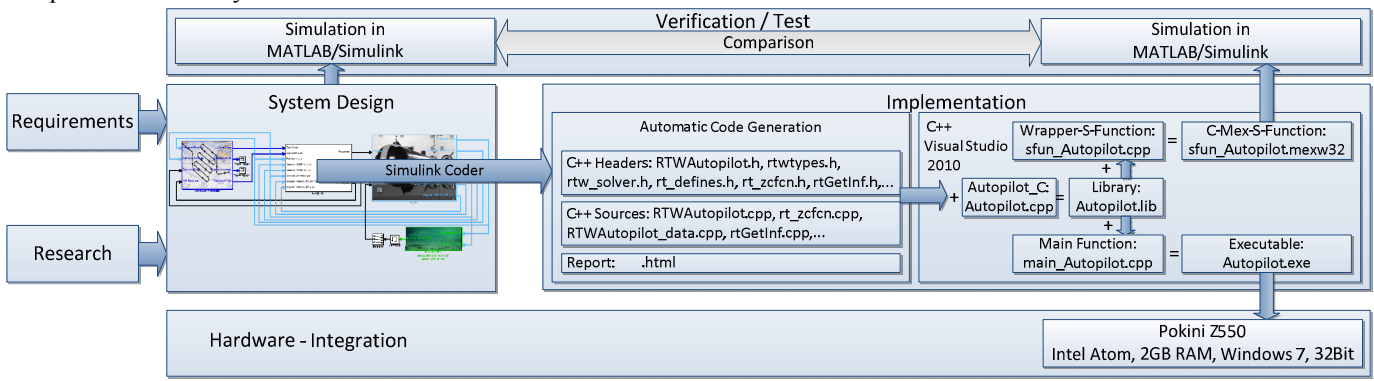

Fig. 8. Model-Based Design of the Autopilot

## A. Wrapper S-Function

The first three test phases use Simulink. To apply a module in Simulink a wrapper S-Function has to be written to include the module code. All modules have a class structure. To create a persistent C++ object of a module in a wrapper S-Function the following steps in the S-Function callback methods are necessary:

1) Create a pointer work vector to store the C++ object:

```
static void mdlInitializeSizes(SimStruct *S)
{
  …
  ssSetNumPWork(S, 1);
  …
}
```

2) Create the object and store the pointer in the pointer work vector:

```
static void mdlStart(SimStruct *S)
{
  void **PWork = ssGetPWork(S);
  …
  Autopilot_C* m_autopilot_p=new Autopilot_C();
  …
  PWork[0] = m_autopilot_p;
}
```

3) Cast the pointer in any S-Function callback method, where it shall be used:

```
static void mdlOutputs(SimStruct *S, int_T tid)
{
  void** PWork = ssGetPWork(S);
  …
  Autopilot_C* m_autopilot_p=(Autopilot_C*)PWork[0];
  …
  m_autopilot_p->ProcessEvent(m_navData_p);
  …
}
```

4) Delete the object when the simulation terminates:

```
static void mdlTerminate(SimStruct *S)
{
  void** PWork = ssGetPWork(S);
  …
  Autopilot_C* m_autopilot_p=(Autopilot_C*)PWork[0];
  …
  delete m_autopilot_p;
  …
}
```

## B. Test phases

### 1) Module Test

The first test phase includes the testing of each single module in Simulink. This test allows for easy verification of the IO behavior of the modules using the available IO blocks of the Simulink Library. It is also possible to write a test script in MATLAB to define complex test sequences and to start the Simulink system.

### 2) Interaction of Modules

In this phase, Simulink systems will be built to test the required behavior between the modules. Fig. 9 shows a Simulink system to verify the interaction between the *Autopilot* and the *Maneuver Processor*.

### 3) Communication Test

To test the different UDP Interfaces the wrapper S-Functions of the modules will use the *UDPInputProcessor_C* and *UDPOutputProcessor_C* instead of the *SimulinkInputProcessor_C* and *SimulinkOutputProcessor_C* Processors (see section IV.B). That means the modules communicate via UDP in Simulink.

### 4) Hardware-in-the-Loop (HIL) Test

All modules run on the SC and communicate via UDP with a vehicle simulator, which runs in Simulink.

### 5) Executable Programs Test

To test all executable programs a simulator will be used, which simulates actuators, sensors and the dynamic behavior of the vehicle CWolf. A socket interface between the simulator and the Virtual Reality CViewVR [17, 18] allows an observation of the vehicle behavior during the simulation. In all programs a message and a data logging function is implemented. This allows for an analysis of all logged messages and data both after the simulation and after a mission.

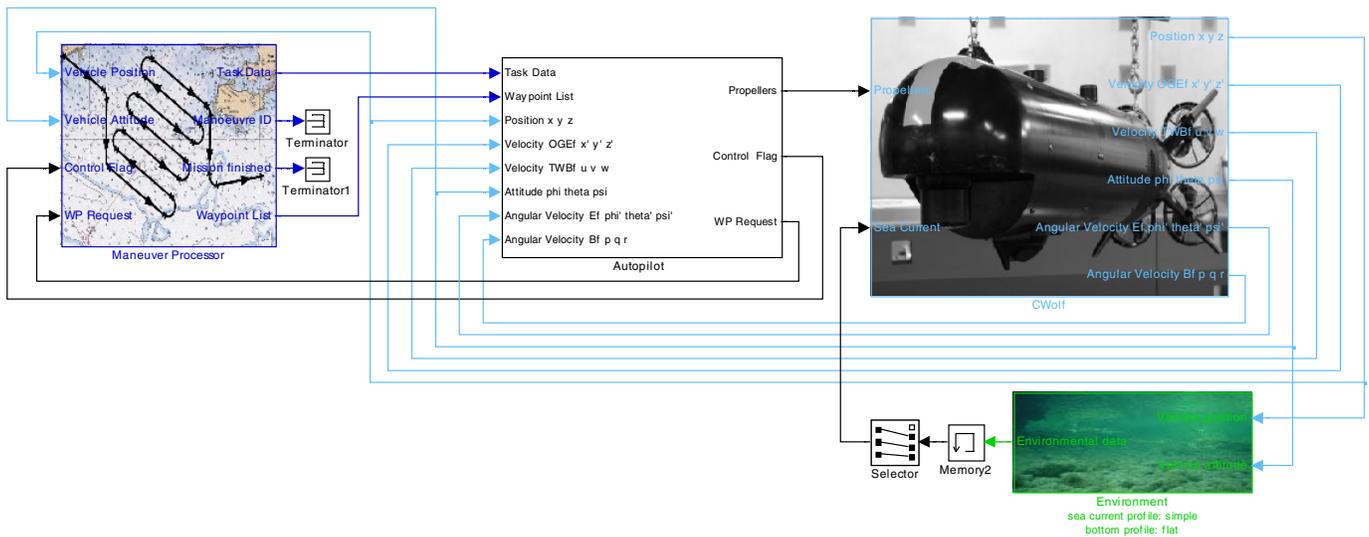

Fig. 9. Simulink System for testing the Autopilot

## VI. CONCLUSIONS AND FUTURE WORK

This paper presents the hard- and software, the mission task and the communication concept of a monitoring system for detection and analysis of water quality parameters and finally a testing platform to achieve high software quality and safe algorithms.

The presented concept allows an easy integration of the payload unit into the vehicle with a minimum of hard- and software technical interfaces. This also allows an easy adaption of other payload units into the vehicle as well as of the payload unit into other vehicle platforms.

The first at-sea trials of the AUV will begin in May 2013. The goal of these tests is to examine the navigation system and to validate the autopilot system on the AUV. It is scheduled to use the monitoring system on fish farms in the Hordaland region/Norway at the end of this year.

## ACKNOWLEDGMENT

We thank Kornelia Bley and Samuel Gebhardt from 4H JENA engineering GmbH for their support and teamwork during the project. Special thanks to our former colleague Dr.-Ing. Matthias Schneider, who designed the easy-to-use mission planning system.